\documentclass{article}
\usepackage{ijcai25}

\usepackage{times}
\usepackage{soul}
\usepackage{url}
\usepackage[hidelinks]{hyperref}
\usepackage[utf8]{inputenc}
\usepackage[small]{caption}
\usepackage{graphicx}
\usepackage{amsmath}
\usepackage{amsthm}
\usepackage{booktabs}
\usepackage{algorithm}
\usepackage{algorithmic}
\usepackage[switch]{lineno}
\usepackage{graphicx}
\usepackage{amsmath}
\usepackage{amssymb}
\usepackage{booktabs}
\usepackage{newfloat}
\usepackage{listings}
\usepackage{subcaption}
\usepackage{caption}
\usepackage{multirow}
\usepackage{balance}
\usepackage{xcolor}
\usepackage{pifont}
\usepackage{colortbl}
\usepackage{adjustbox}
\usepackage{array, makecell}
\usepackage{tikz}
\usepackage{tabularx}
\definecolor{lightblue}{rgb}{0.68, 0.85, 0.9}
\definecolor{lightpurple}{rgb}{0.85, 0.7, 0.85} 
\definecolor{lightgreen}{RGB}{173, 255, 173}
\definecolor{lightorange}{RGB}{255, 204, 153}

\usepackage{adjustbox}

\title{VRD-IU: Lessons from Visually Rich Document Intelligence and Understanding}

\author{
Yihao Ding$^{1,2}$
\and
Soyeon Caren Han$^{1,2}$\and
Yan Li$^2$\And
Josiah Poon$^2$\\
\affiliations
$^1$The University of Melbourne,
$^2$The University of Sydney\\
\emails
\{yihao.ding, josiah.poon\}@sydney.edu.au,
yali3816@uni.sydney.edu.au
caren.han@unimelb.edu.au
}

\begin{document}

\maketitle

\begin{abstract}
Visually Rich Document Understanding (VRDU) has emerged as a critical field in document intelligence, enabling automated extraction of key information from complex documents across domains such as medical, financial, and educational applications. However, form-like documents pose unique challenges due to their complex layouts, multi-stakeholder involvement, and high structural variability. Addressing these issues, the VRD-IU Competition was introduced, focusing on extracting and localizing key information from multi-format forms within the Form-NLU dataset, which includes digital, printed, and handwritten documents.
This paper presents insights from the competition, which featured two tracks: Track A, emphasizing entity-based key information retrieval, and Track B, targeting end-to-end key information localization from raw document images. With over 20 participating teams, the competition showcased various state-of-the-art methodologies, including hierarchical decomposition, transformer-based retrieval, multimodal feature fusion, and advanced object detection techniques. The top-performing models set new benchmarks in VRDU, providing valuable insights into document intelligence.
\end{abstract}

\section{Introduction}
Visually Rich Document Understanding (VRDU) has garnered significant attention across various domains, including medical \cite{pdfmvqa}, receipts \cite{cord,sroie}, financial \cite{nda}, and educational \cite{vies} fields, becoming a pivotal approach for recording, preserving, and sharing information. Recent advancements in deep learning, particularly self-supervised pretrained transformers \cite{layoutlmv3,udop,layoutxlm} and large language models \cite{blip3,gpt4o,llava}, have demonstrated notable success in this domain.
However, form-like documents present unique challenges due to their involvement of multiple parties (designers, deliverers, and fillers), complex layouts, and higher uncertainties. These factors create significant barriers to directly applying existing methods to practical tasks \cite{m3vrd,ding2024david}.

\begin{figure}[tb]
    \hspace*{0cm}
     \centering
     \hspace*{0em}
     \begin{subfigure}[b]{0.15\textwidth}
         \centering
         \includegraphics[height=3.3cm]{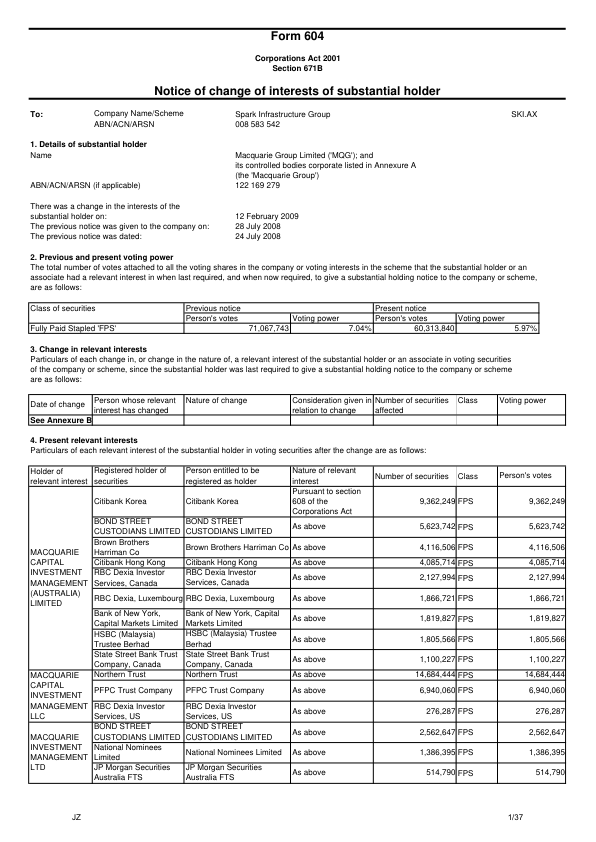}
         \caption{Digital}
         \label{fig:digital_sample}
     \end{subfigure}
     \hspace*{0em}
     \begin{subfigure}[b]{0.15\textwidth}
         \centering
         \includegraphics[height=3.3cm]{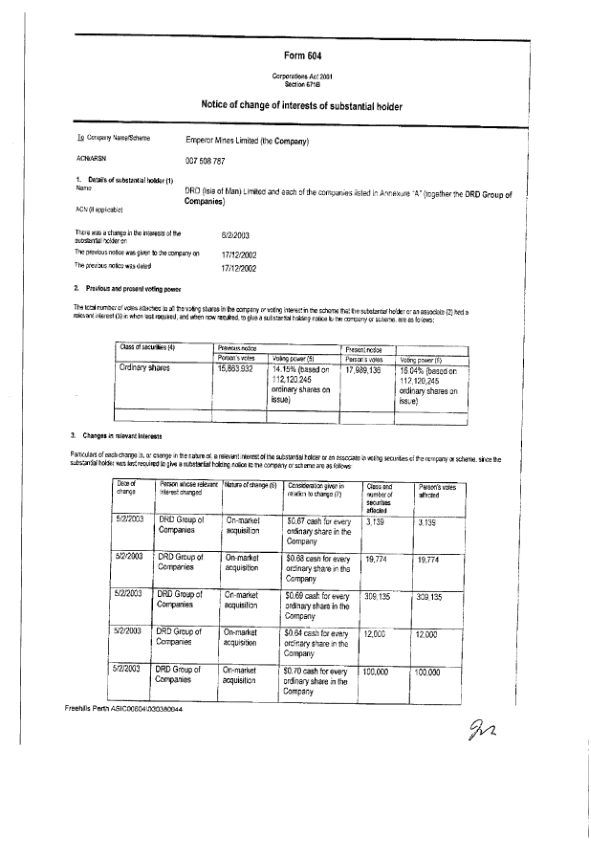}
         \caption{Printed}
         \label{fig:printed_sample}
     \end{subfigure}
     \hspace*{0em}
     \begin{subfigure}[b]{0.15\textwidth}
         \centering
         \includegraphics[height=3.3cm]{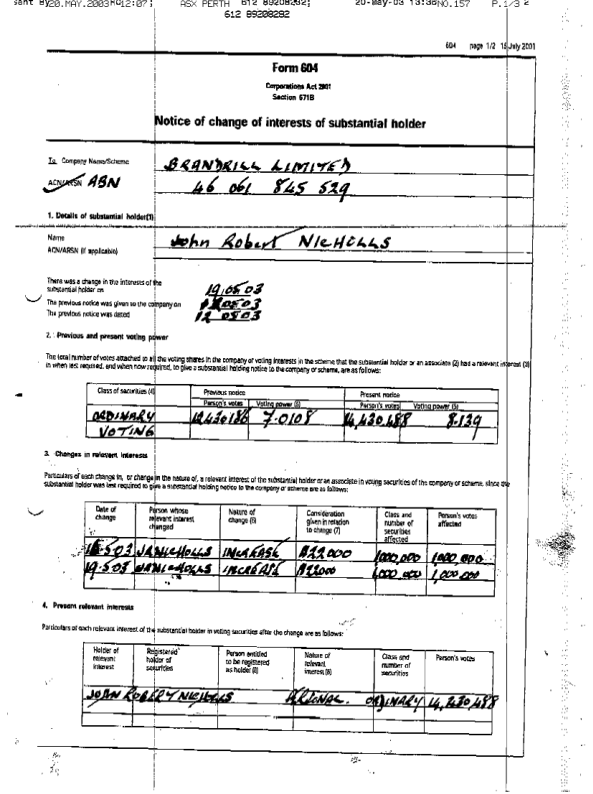}
         \caption{Handwritten}
         \label{fig:handwritten_sample}
     \end{subfigure}
        \caption{Digital, Printed and Handwritten Form Samples from Form-NLU dataset.}
        \label{fig:form_samples}
            \vspace{-1em}
\end{figure}

To address the challenges caused by form-like documents, we introduce the VRD-IU Competition, which focuses on extracting key information from multi-format forms within the Form-NLU dataset \cite{formnlu}, encompassing digital, printed, and handwritten documents, as illustrated in Figure~\ref{fig:form_samples}. The competition aims to tackle the challenges associated with the diverse and complex nature of such documents, which often involve multiple stakeholders and contain essential information that is difficult to extract. By offering two distinct tracks tailored to different skill levels, it accommodates participants ranging from newcomers to seasoned professionals in the field of deep learning. This initiative not only fosters innovation in document understanding and accelerates advancements in information extraction techniques but also aims to engage a broader community, encouraging innovators to contribute to the evolution of efficient information analysis methodologies. 

More than 20 teams participated in the VRD-IU competition, with the top 3 teams in Track A and the top 2 teams in Track B submitting their code and abstract papers to share their proposed solutions. Various methods were proposed, focusing on achieving more comprehensive document understanding. In Track A, the methods mainly focused on acquiring more representative entity features, while in Track B, end-to-end frameworks were proposed to enable the framework to understand both the semantic and structural aspects of form images to extract and locate the relevant information. The proposed methods achieved new state-of-the-art performance and paved the way for future work in complex form understanding direction.

The contributions of this paper are as follows: 

1) To provide a redefined task definition for the Form-NLU dataset, focusing on Key Information Localization (Track-B), which involves both extracting and localizing target information from input form images. 

2) To introduce the state-of-the-art approaches proposed by competition participants, along with a qualitative analysis of these methods. 

3) To offer key insights into future directions for VRDU understanding, highlighting potential areas for improvement and innovation.

\section{VRD-IU Competition Task Definition}
The adopted dataset for the VRD-IU competition is derived from the Form-NLU dataset. Unlike the original tasks, which emphasize both form structure and content understanding, this competition specifically targets the extraction of key information from complex form documents. To address varying levels of task complexity, two distinct tracks are introduced. Track A aligns with Form-NLU Task B, focusing on retrieving key information from provided semantic entities within the document. In contrast, Track B presents a more challenging scenario, offering only the form image and query without any prior information about the document structure. The detailed definitions of these tracks are provided to support future research in this area.
\begin{figure}[ht]
  \centering
   \includegraphics[width=\linewidth]{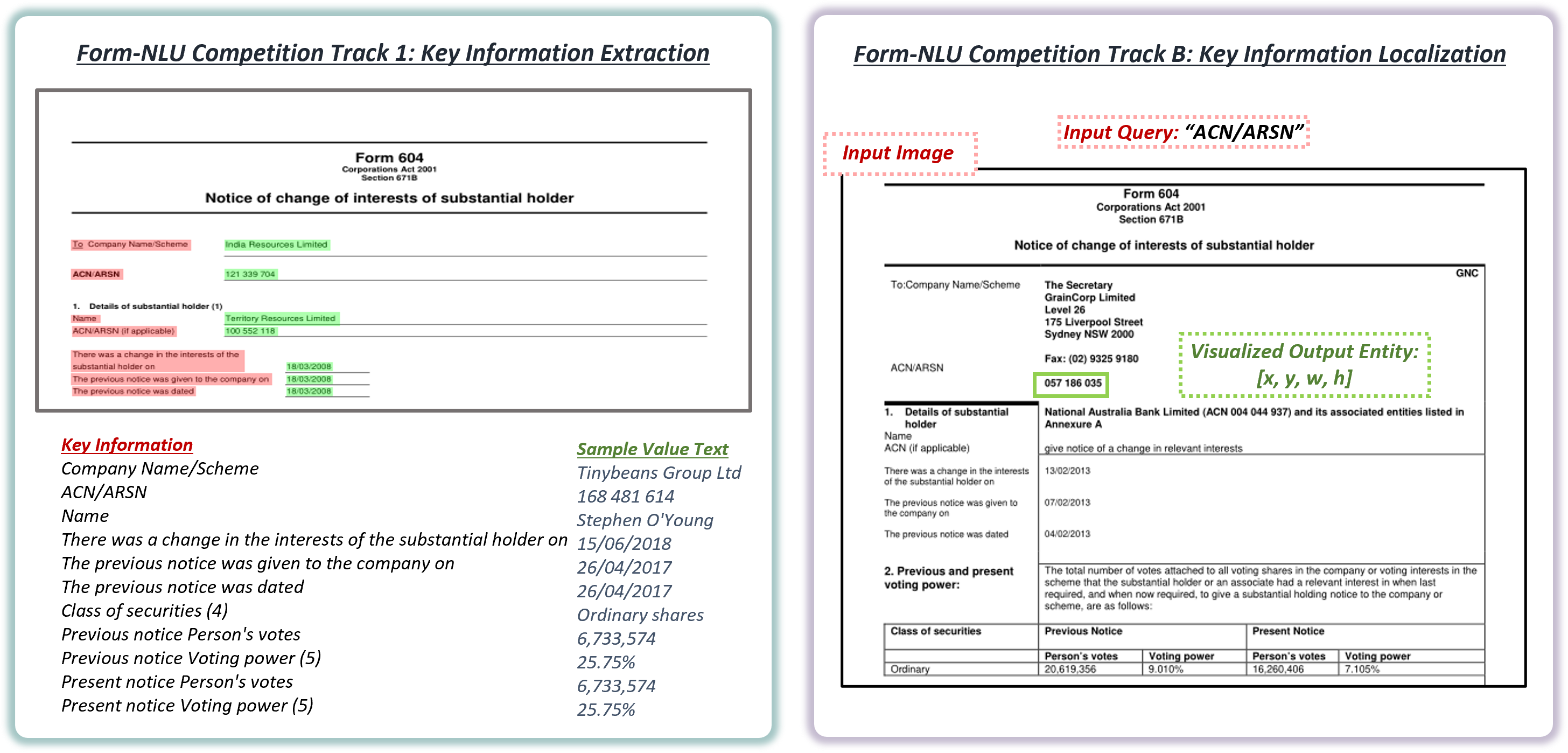}
  \caption{VRD-IU Competition Track A and Track B Samples.}
  \label{fig:task_definition}
\end{figure}

\noindent \textbf{Track A: Form Key Information Extraction} \footnote{The dataset used for Track A can be found via \url{https://www.kaggle.com/competitions/vrd-iu2024-tracka}}
Track A involves developing deep learning-based retrieval models to extract specific form components in response to a given key query. Human-annotated bounding box coordinates representing the semantic entities of input form documents are provided. The task requires accurate identification and localization of the target entity based on the query. Performance is evaluated using the \textbf{F1-Score}, following the evaluation methodology defined in Form-NLU Task B.

\noindent\textbf{Track B: Form Key Information Localization} \footnote{The dataset used for Track B can be found via \url{https://www.kaggle.com/competitions/vrd-iu2024-trackb}}
Track B focuses on designing an end-to-end framework to predict bounding box coordinates of form components directly from input document images based on a given key query. Unlike Track A, ground truth bounding boxes for semantic entities are not provided; the inputs consist solely of form images and query keys. Evaluation is based on the \textbf{Mean Average Precision (MAP)} of the predicted bounding boxes, reflecting both accuracy and localization effectiveness.

\section{Methods}
\subsection{Methods for Track-A}
To effectively retrieve key information from domain-specific semantic entities based on user queries, several solutions leverage enhanced document understanding through pretrained transformers. These approaches incorporate advanced components and strategies hierarchical decomposition, token classification, and multimodal feature fusion, to better comprehend the structure and semantics of forms, enabling more accurate and efficient extraction of target information. 

\noindent \textbf{Key Information Extraction with Hierarchical Layout Decomposition}, proposed by Choi et al. (\textbf{Team rb-ai}, \textit{Top-1 Team in Track-A}), utilizes prior template knowledge from the Form-NLU dataset to segment form images into top, middle, and bottom sections. Then, section-specific KIE models are then applied for key information extraction. To address misclassification issues within each section, heuristic post-processing techniques are employed. The approach uses a pretrained transformer-based YOLOv8 as an object detector for section segmentation, while LayoutLMv3 \cite{layoutlmv3} is fine-tuned as the KIE backbone for each segmented section.

\noindent \textbf{Redefining Information Extraction from Visually Rich Documents as Token
Classification} is introduced by Song et.al (\textbf{Team vipski}, \textit{Top-2 Team in Track A}) to convert the entity retrieving task as a sequence labelling task with maintaining the original resolution and aspect ratios of document images, the approach reduces visual distortion, enhancing model accuracy. After fine-tuning LayoutLM-v3 with a token classifier, the proposed method achieve cutting-edge performance. 

\noindent \textbf{Transformer-Based Visual Feature and Textual Feature Fusion Model
for Document Key Information Extraction}, proposed by Kim et al. (\textbf{Team gcu}, \textit{Top-3 Team in Track A}), aims to enhance document semantic entity representation. RoI visual and textual features are separately processed through a transformer-based vision-language encoder. A cross-encoder then strengthens the correlations between questions and entities. The combined textual and visual representations are fed into a classifier for final information retrieval.

\subsection{Methods for Track B}
Different from Track A providing the bounding box coordinate of each document semantic entity, Track B requires the model can detect the exact object (semantic entities) from the document image. To realize this, except for the well-designed object detection approaches, some domain-adaptation strategies are adopted.

\noindent \textbf{Key Information LocalizationWith Hierarchical Ensemble} is introduced for key information localization by Choi et al. (\textbf{Team rb-ai}, \textit{Top-1 Team in Track-A}). The approach uses YOLO \cite{yolo} and RT-DETR \cite{rtdert} models as baselines and integrates their predictions through a global ensemble and a layout-specific ensemble. Additionally, a layout-based post-editor refines predictions using text recognition and layout-specific detectors to correct mispredictions.

\noindent \textbf{Enhancing Document Key Information Localization Through Data Augmentation} is proposed by Dai et al. (\textbf{Team chatdy}, \textit{Top-3 Team in Track-A}), which divides the whole workflow to document augmentation phase and an object detection phase. The augmentation phase uses the Augraphy library to mimic the appearance of handwritten documents from digital ones, emplointoying techniques like InkBleed, Letterpress, and JPEG compression to simulate handwritten characteristics. The object detection phase utilizes pretrained VRDU models like DiT \cite{dit} and LayoutLMv3 with Faster R-CNN \cite{fasterrcnn} and Mask R-CNN \cite{maskrcnn} frameworks. The key innovation lies in the augmentation strategy, which enhances the model's ability to generalize to handwritten documents, even when trained only on digital data.
\section{Competition Results}
The VRD-IU competition attracted significant participation from teams employing diverse methodologies to tackle the key information extraction and localization challenges. Track A and B's evaluation results reveal notable trends in model performance and methodological efficacy.

\noindent \textbf{Track A: Form Key Information Extraction}\\
The results of the top-performing teams in Track A are shown in Table \ref{tab:competition-results-tracka}. The results demonstrate that pretrained transformer-based approaches, enhanced with hierarchical decomposition and multimodal feature fusion, are highly effective in extracting semantic entities from form documents.

\noindent \textbf{Track B: Form Key Information Localization}\\
Performance in Track B was evaluated using Mean Average Precision (MAP), reflecting both detection accuracy and localization effectiveness, shown in Table \ref{tab:competition-results-trackb}.
Track B posed a greater challenge due to the lack of predefined bounding box annotations. The results indicate a clear performance gap compared to Track A, underscoring the complexity of end-to-end document understanding directly from form images.

\begin{table}[ht]
\centering
\begin{adjustbox}{width=\linewidth}
\begin{tabular}{clcccc}
\toprule
\textbf{Rank} & \textbf{Team Name} & \textbf{Members} & \textbf{Private (\%)} & \textbf{Public (\%)}  \\ 
\midrule
1 & \textbf{rb-ai} & 3 & \textbf{100} & \textbf{100}  \\ 
2 & \textbf{vipski} & 2 & 97.93 & \textbf{100}  \\ 
3 & \textbf{gcu} & 4 & 96.95 & 99.67 \\ 

4 & MKAZ & 1 & 93.07 & 99.57 \\ 
5 & chatdy & 1 & 92.57 & 99.57  \\ 
6 & Play4fun & 1 & 90.04 & 99.08  \\ 
7 & Team G.AI & 3 & 80.17 & 97.78  \\ 
8 & dagmbari & 1 & 78.99 & 95.55  \\ 
9 & jaehongleee & 1 & 78.87 & 93.74 \\ 
10 & zuo-zou & 1 & 48.45 & 80.09 \\ 
11 & DualMonitor & 2 & 42.23 & 64.13\\ 
12 & rlawnsxo & 1 & 41.45 & 54.84 \\ 
\bottomrule
\end{tabular}
\end{adjustbox}
\caption{Track A Key Information Extraction Results.}
\label{tab:competition-results-tracka}
\end{table}

\begin{table}[ht]
\centering
\begin{adjustbox}{width=\linewidth}
\begin{tabular}{clcccc}
\toprule
\textbf{Rank} & \textbf{Team Name} & \textbf{Members} & \textbf{Private (\%)} & \textbf{Public (\%)}  \\ 
\midrule
1 & \textbf{rb-ai} & 4 & \textbf{65.79} & \textbf{99.70} \\ 
2 & \textbf{oaths11} & 3 & 56.97 & 80.97 \\ 
3 & \textbf{chatdy} & 2 & 55.68 & 80.49 \\ 
4 & gcu & 1 & 43.93 & 79.18 \\ 
5 & BiaoXup & 1 & 18.12 & 13.63 \\ 
6 & Shuo Yang & 1 & 17.06 & 13.03 \\ 
7 & VRD IU & 1 & 16.80 & 11.66 \\ 
8 & eunyi lyou & 1 & 0.23 & 4.88 \\ 
\bottomrule
\end{tabular}
\end{adjustbox}
\caption{Track B Key Information Localisation Results}
\label{tab:competition-results-trackb}
\vspace{-2em}
\end{table}

\section{Key Insights}
The competition provided valuable insights into document understanding tasks, particularly within visually rich document (VRD) contexts:

\noindent \textbf{1. Hierarchical Decomposition and Post-Processing Improve Accuracy} \
Track A results highlighted that segmenting form structures into hierarchical components significantly enhances extraction accuracy. Techniques such as section-wise key information extraction (as seen in rb-ai’s approach) help mitigate structural misclassification and boost precision.

\noindent \textbf{2. Pretrained Transformers Excel in Structured Document Tasks} \
Methods leveraging LayoutLMv3 and other pretrained vision-language transformers outperformed conventional models, reaffirming the effectiveness of multimodal pretraining for VRD tasks. Fine-tuning on domain-specific data further improved entity extraction capabilities.

\noindent \textbf{3. Object Detection Remains a Challenge for Form Localization} \
Track B results highlighted the difficulty of localizing key information without predefined bounding boxes. Despite the effectiveness of YOLO and RT-DETR models, the lower MAP scores indicate that further advances in end-to-end document understanding are needed.

\noindent \textbf{4. Data Augmentation Enhances Generalization} \
Augmenting digital forms with synthetic handwritten effects improved model robustness for handwritten document retrieval (e.g., ChatDY’s approach). This suggests that document variability must be carefully considered in future dataset preparation.

\noindent \textbf{5. Ensemble Approaches Provide Performance Gains} \
Models integrating multiple object detection techniques (such as rb-ai’s hierarchical ensemble strategy) showed superior performance, indicating that fusing diverse model outputs can enhance accuracy in complex layouts.

\section{Conclusion}
The VRD-IU competition served as an effective benchmark for evaluating state-of-the-art approaches in key information extraction and localization within form documents. Overall, the VRD-IU competition has accelerated research in visually rich document understanding by providing a well-defined benchmark and fostering innovation in key information extraction and localization. The insights gained will be instrumental in guiding the development of more robust, efficient, and generalizable AI-driven document processing systems.

\appendix

\section*{Ethical Statement}

There are no ethical issues.

\section*{Acknowledgments}
Thanks for the support from Google Research, United States.

\section*{Contributors}
We would like to acknowledge the invaluable contributions of the authors across all competition winners in IJCAI 2024 VRD-IU Competition:
\begin{itemize}

    \item \textbf{Track A - Team rb-ai (Rank 1): Key Information Extraction with Hierarchical Layout Decomposition}: \textbf{Hongjun Choi} (\textit{Teamreboott Inc., South Korea}), \textbf{Jongho Lee} (\textit{Teamreboott Inc., South Korea}; \textit{Pukyong National University, South Korea}), and \textbf{Jaeyoung Kim} (\textit{Teamreboott Inc., South Korea}).
    
    \item \textbf{Track A - Team vipski (Rank 2): Redefining Information Extraction from Visually Rich Documents as Token Classification}: \textbf{Jonghyun Song} and \textbf{Eunyi Lyou} (\textit{Graduate School of Data Science, Seoul National University}).

    \item \textbf{Track A - Team gcu (Rank 3): Transformer-Based Visual Feature and Textual Feature Fusion Model for Document Key Information Extraction}: \textbf{Wooseok Kim}, \textbf{Juhyeong Kim}, \textbf{Sangyeon Yu}, \textbf{Gyunyeop Kim}, and \textbf{Sangwoo Kang} (\textit{School of Computing, Gachon University}).

    \item \textbf{Track B - Team rb-ai (Rank 1):Key Information Localization With Hierarchical Ensemble}: \textbf{Hongjun Choi} (\textit{Teamreboott Inc., South Korea}), \textbf{Jongho Lee} (\textit{Pukyong National University, South Korea}), and \textbf{Jaeyoung Kim} (\textit{Teamreboott Inc., South Korea}).
    
    \item \textbf{Track B - Team chatdy (Rank 2):Enhancing Document Key Information Localization Through Data Augmentation}: \textbf{Yue Dai} (\textit{The University of Western Australia}).
\end{itemize}

Their dedication and innovative research have significantly advanced the field of key information extraction and visually rich document understanding.
\bibliographystyle{named}
\bibliography{ijcai25}

\end{document}